\documentclass{article}

\usepackage[nonatbib, final]{neurips_2019}

\usepackage[utf8]{inputenc} % allow utf-8 input
\usepackage[T1]{fontenc}    % use 8-bit T1 fonts
\usepackage{hyperref}       % hyperlinks
\usepackage{url}            % simple URL typesetting
\usepackage{booktabs}       % professional-quality tables
\usepackage{amsfonts}       % blackboard math symbols
\usepackage{nicefrac}       % compact symbols for 1/2, etc.
\usepackage{microtype}      % microtypography

\usepackage{times}
\usepackage{epsfig}
\usepackage{graphicx}
\usepackage{amsmath}
\usepackage{amssymb}
\usepackage{multirow}
\usepackage{xspace}
\usepackage{xcolor}
\usepackage{floatrow}
\usepackage{subcaption} 
\newfloatcommand{capbtabbox}{table}[][\FBwidth]
\usepackage{float}
\newcommand{\specialcell}[2][c]{%
  \begin{tabular}[#1]{@{}c@{}}#2\end{tabular}}

\restylefloat*{figure}

\usepackage{blindtext}

\newcommand*{\eg}{e.g.\@\xspace}

\title{Only Time Can Tell: \\ Discovering Temporal Data for Temporal Modeling}

\author{%
  Laura Sevilla-Lara\thanks{Part of the work was done while the author was at Facebook AI.}\\
  University of Edinburgh\\
   \And
   Shengxin Zha \\
   Facebook AI \\
   \AND
   Zhicheng Yan \\
   Facebook AI \\
   \And
   Vedanuj Goswami \\
   Facebook AI \\
   \And 
   Matt Feiszli \\ 
   Facebook AI \\
   \And
   Lorenzo Torresani \\
   Facebook AI \\
}

\begin{document}

\maketitle

\begin{abstract}

Understanding temporal information and how the visual world changes over time, is a fundamental ability of intelligent systems. In video understanding, temporal information is at the core of many current challenges, including compression, efficient inference, motion estimation or summarization. However, in current video datasets it has been observed that action classes can often be recognized without any temporal information, from a single frame of video. As a result, both benchmarking and training in these datasets may give an unintentional advantage to models with strong image understanding capabilities, as opposed to those with strong temporal understanding. In other words, current datasets may not reward good temporal understanding, potentially hindering progress.   
In this paper we address this problem head on by identifying action classes where temporal information is actually necessary to recognize them and call these ``temporal classes''. Selecting temporal classes using a computational method would bias the process. Instead, we propose a methodology based on a simple and effective human annotation experiment. We remove just the temporal information, by shuffling frames in time, and measure if the action can still be recognized. Classes that cannot be recognized when frames are not in order, are included in the {\em Temporal Dataset}. We observe that this set is statistically different from other static classes, and that performance in it correlates with a network's ability to capture temporal information. Thus we use it as a benchmark on current popular networks, which reveals a series of interesting facts, like inflated convolutions bias networks towards classes where motion is not important. We also explore the effect of training on the temporal dataset, and observe that this leads to better generalization in unseen classes, demonstrating the need for more temporal data. We hope that the proposed dataset of temporal categories will help guide future research in temporal modeling for better video understanding.% \ls{revise these last few sentences.}

% We also observe that training on these classes produces stronger temporal features, which reveal a series of interesting facts, such as that 3D convolutional networks trained on this subset of classes have stronger generalization than those trained on a larger, unfiltered dataset. 

%We present computational analyses on this set of temporal classes, which reveal a series of interesting facts, such as that 3D convolutional networks trained on this subset of classes have stronger generalization than those trained on a larger, unfiltered dataset. We hope that our proposed ``meta-dataset" of temporal categories will help guide future research in temporal modeling for better video understanding. %\girum{clear concept with a solid motivation behind it.}
\end{abstract}

\begin{figure}[h]
    \centering
    \includegraphics[width=\textwidth]{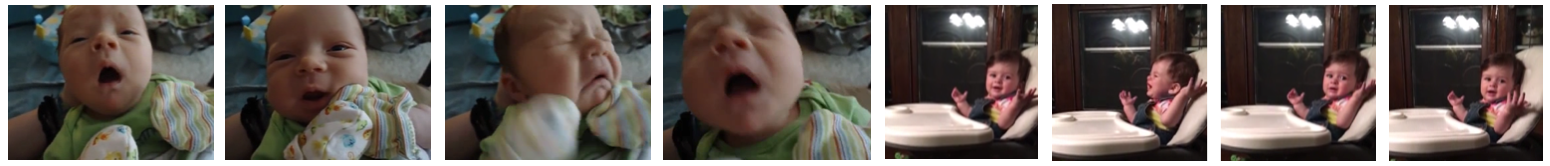}
    \caption{{\bf Can you guess these actions? ``yawning", ``sneezing" or ``crying"? } 
    Temporal information is essential to discriminate some actions, while for others it is redundant. 
    Shuffling frames in time removes temporal information, revealing the actions where it actually matters.  
    (Solution at the end of the paper.) }
    %classes can be recognized only after we hit the play button. }
\end{figure}%

%%%%%%%%% BODY TEXT
\section{Introduction}

\begin{figure}
  \begin{subfigure}[b]{0.48\textwidth}
        \centering
        \includegraphics[width=0.95\textwidth]{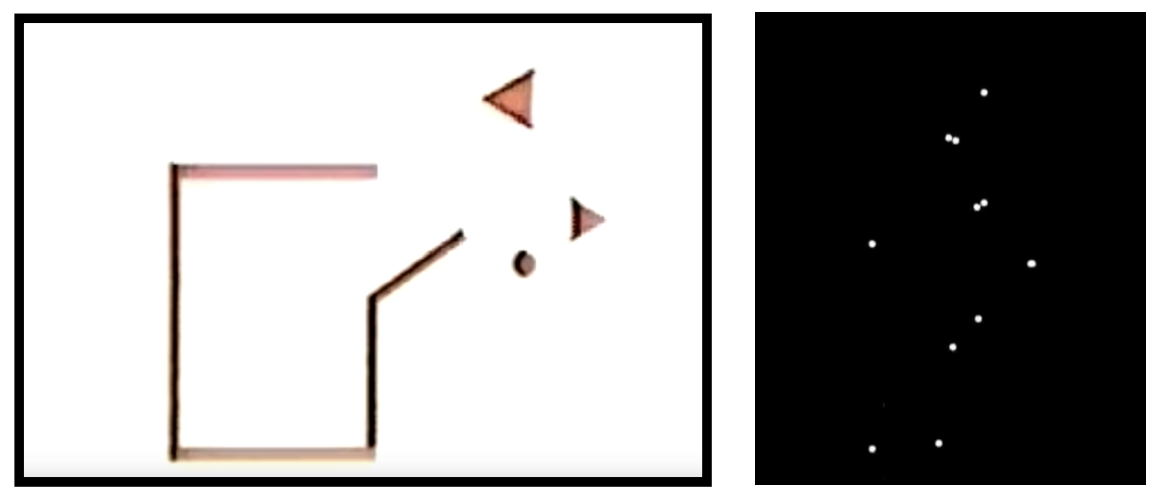}
        \label{fig:heider_simmel}
        \caption{{Temporal task.} }
    \end{subfigure}\hfill
    \begin{subfigure}[b]{0.46\textwidth}
        \centering
        \includegraphics[width=\linewidth]{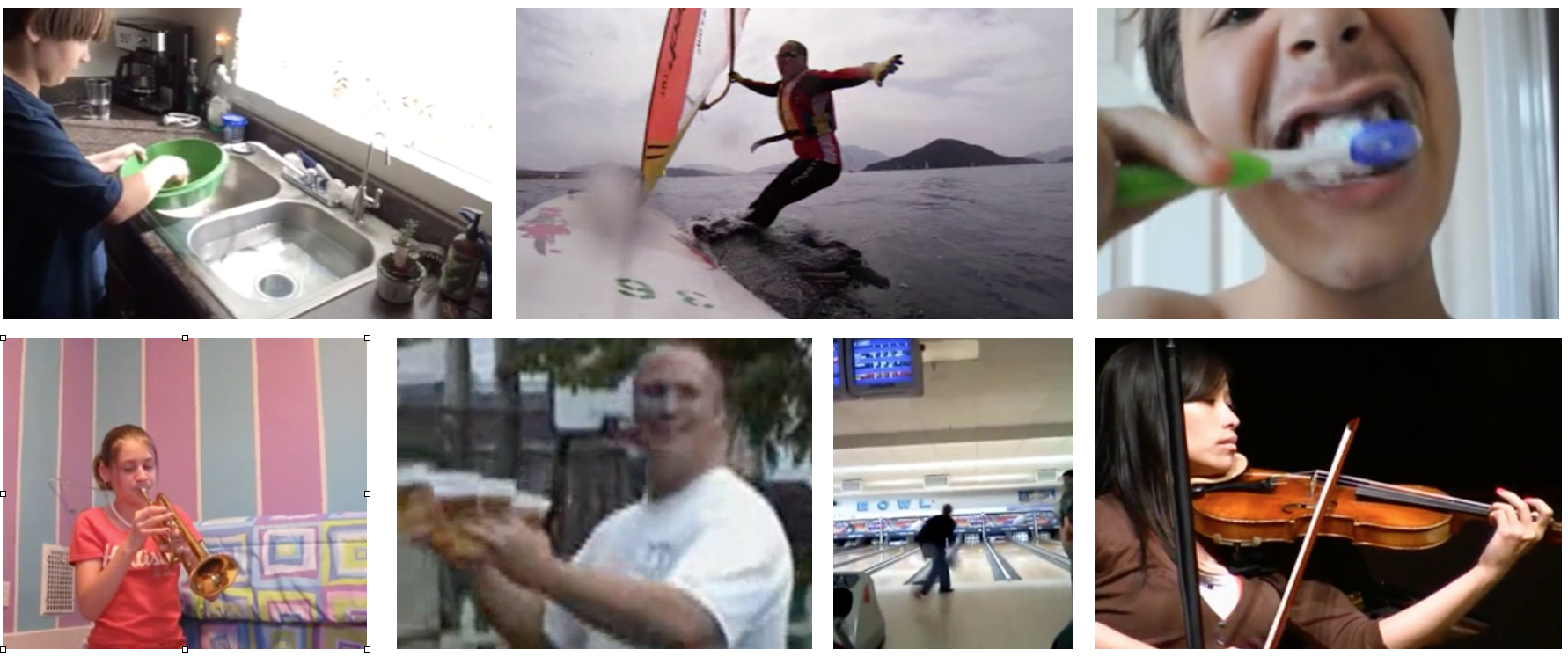}
        \caption{{Static task.} }
    \end{subfigure}
    \caption{{\bf The relevance of temporal information varies vastly by task. } In one end of the spectrum (a) shows sample frames from Heider and Simmel~\cite{heiderandsimmel} (left) and Johansson~\cite{johansson1973} (right). Images alone do not give much information, but when the video is played, an action or a even a story comes to life. These experiments highlight the richness of temporal information in visual understanding. At the other end (b) shows sample frames from Kinetics. It is easy to map images to {\em playing trumpet, brushing teeth, washing dishes, drinking beer, playing violin, windsurfing, bowling}. In this task, temporal information is completely redundant. }
    \label{fig:frame_samples}
\end{figure}

%~\ls{Teaser figure?? maybe videos with shuffled frames that are difficult to understand? }
%-- Motion has shown to be important for recognition in the biological setting. (Johansson, 40s experiment)
Temporal information has been shown to be important for recognition. In the biological setting, Johansson's point-light experiments~\cite{johansson1973} show that moving dots alone contain enough information to understand actions, when the points-lights are placed at the joints. Similarly, Heider and Simmel~\cite{heiderandsimmel} find that simple figures (like squares or triangles) in motion can convey rich story information, including intention and emotions like anger, joy. Figure~\ref{fig:frame_samples}(a) shows two frames from these experiments; neither the action nor a story can be recognized easily in the absence of temporal information.

%-- It has also shown to be important for recognition in a computational setting. (UCF101, HMDB51) 
In computational settings, temporal features (optical flow, spatio-temporal convolutions, or both) have been shown to be useful for video understanding.  For some datasets and models, using optical flow as input actually yields higher accuracy than using RGB images~\cite{zissermantwostream, i3d, r2plus1d}.
%\girum{the figure on the first page is very descriptive. I think it will be even better with a subcaption, assuming those are frames from two different videos.}
% -- However, we observe that in many datasets single images are very informative. 
%-- also the larger datasets this is not the case: flow contains a very small amount of complementary information. 
However, a single image is often informative enough to predict the label with good confidence. %Hence we pose the question: 
This leads to the question {\em when do we need temporal information to understand an action?} %\lt{I rephrased the question in the form of {\em when} as this seems to better capture the goal of the paper.}
Figure~\ref{fig:frame_samples}(b) shows some examples from Kinetics~\cite{kinetics}, where the action class is easy to guess. 
%At the same time it is interesting to notice that on this same dataset, two-stream models such as I3D~\cite{i3d}, achieve higher accuracy with the stream operating on optical flow (the temporal stream) compared to the stream using RGB frames as input (the spatial stream).  
Furthermore, in recent two-stream models such as I3D~\cite{i3d} fusing the predictions of the stream operating on optical flow (the temporal stream) to that operating on RGB frames (the spatial stream) produces rather small gains. %This suggests that for certain classes or specific videos in the Kinetics dataset, temporal analysis remains critical for accurate computational recognition, while appearance information adds only marginally to performance.
%-- We also observe (Dhruv) that a single frame, or shuffled frames have a small effect on recognition. 
%\ls{mention Dhruv's paper too?} 
%It is thus not surprising that much of the progress on video classification methods has focused on adapting image recognition strategies to video. As a result, the 
%-- Does this mean that motion is not important?
%-- We argue that these observations do not reflect the importance of motion but rather the choice of categories in current video datasets. Images alone are often enough to predict many of these categories. 
%Inspired by these observations, we argue that in most existing datasets, certain action classes benefit more than others from motion analysis for recognition. 
Do these observations mean that given the image content, motion information is redundant? We argue that actually these observations do not reflect the importance of motion in video understanding, but rather the distribution of categories in some of the video datasets. %In such categories, images alone are indeed often enough to predict many of these categories. 
Based on this hypothesis, we propose to examine some of the most widely used datasets, and identify classes where temporal information is necessary to recognize them. We call these ``temporal classes." Our analysis is based on human perception rather than on the performance of computational models as we do not want to tie the definition of temporal categories to a specific model's performance. %Here we use the human visual system as a baseline model of temporal relevance for action categorization. 
In particular, given a collection of videos , we compare human performance on a video with shuffled frames and on the same video with the original ordering of frames. We use this difference as a cue for how necessary temporal information is to recognize a class. We then select the classes where this difference is largest. 
%We identify the categories where recognition accuracy degrades most, which  temporal analysis for good recognition. In this way we can determine how often temporal classes occur in some widely used video datasets.
We use these categories as a benchmark to measure how well different architectures perform. Our analyses provides interesting observations such as that the ranking of models on the temporal classes is different compared to that on the full datasets and that temporal models trained on temporal classes have stronger generalization performance. %We also use these categories to train ~\ls{fill in with observations}. 
% -- Given this collection of categories, analyze what families of categories require motion. 
% -- We also use this as benchmark for how well different architecture choices perform. 
% -- We also train on these categories... 
% -- Why can we not just shuffle images? because we cannot disambiguate if it is the lack of ability of temporal modeling or the 
In summary, we make the following contributions: 
\begin{itemize}
    \item A {\em methodology to discover the importance of temporal information} in action classes based on human perception. This allows us to untie the definition of temporal relevance from the performance of existing computational methods. 
    \item The {\em Temporal Dataset, resulting from using our proposed methodology on current video classification datasets}. We identify the set of categories containing more temporal information and what they have in common, and examine their computational properties. We will release all of the collected human judgments for further analysis by the community. 
    \item We {\em benchmark current video models}, and observe that the ranking changes with the inclusion or omission of temporal categories, revealing hidden biases and strengths. This also suggests that the distribution of the videos in the temporal categories actually has different structure. %We also observe that the motion categories not only are different, but they also contain more temporal information; in particular, shuffling at test time affects these categories most. 
    %\girum{maybe it is better to name them as  motion-dependent (non-exchangeable) and motion-independent(exchangeable) actions for clarity}.  
    \item We use the {\em temporal categories for training temporal models}. We observe that training on the Temporal Dataset produces stronger temporal features. We also observe that training on this dataset leads to a better ability to generalize in the transfer learning setting of training on ``seen" categories and testing on ``unseen" categories. 
\end{itemize}
We hope that our proposed ``meta-dataset" of temporal categories will help guide future research in temporal modeling for better video understanding. 

% ~\footnote{\url{https://www.youtube.com/watch?v=rEVB6kW9p6k&t=60s}}
% ~\footnote{\url{https://www.youtube.com/results?search_query=heider+and+simmel}}.
%We hope that these temporal categories can become a good benchmark for future research on temporal modeling. 

% \begin{figure}
% \centering
% \includegraphics[width=\linewidth]{figures/heider_simmel_johansson3_small.png}
% \caption{{\bf Single frames from the experiments of Heider and Simmel~\cite{heiderandsimmel} (left) and Johansson~\cite{johansson1973} (right).} These experiments highlight the importance of temporal information in video understanding. In these examples, single images do not convey much information, the story is difficult to understand. However, if we play the videos we can easily recognize a human performing different actions in the case of Johansson's experiment and a rich story in the case of Heider and Simmel's study.}
% \label{fig:heider_simmel}
% %\vspace{-0.6cm}
% \end{figure}

% \begin{figure}
% \centering
% \includegraphics[width=\linewidth]{figures/kinetics_easy_small.png}
% \caption{{\bf Sample frames from Kinetics.} Can you tell which image corresponds to the following classes? {\em playing trumpet, brushing teeth, washing dishes, drinking beer, playing violin, windsurfing, bowling}}
% \label{fig:kinetics_samples}
% \end{figure}

\section{Related Work}

\noindent \textbf{Temporal Modeling}. Over the last years video understanding has made rapid progress. While modeling spatial information on videos has largely been addressed leveraging advancements in image understanding methods, modeling temporal information has sprouted new and creative techniques. These techniques can be broadly categorized in three families: two-stream, temporal convolutions, and recurrent methods. The first, pioneered by Simonyan and Zisserman~\cite{zissermantwostream}, consists on having two parallel networks: the spatial stream aimed at modeling image content, which takes RGB as input, and a temporal stream aimed at modeling motion content, which takes a small number of neighboring flow fields as input.  The predictions of the two streams are combined at the end. Different variants of this general framework arise using different architectures for the streams~\cite{tsn}. These techniques are described as frame-based, since they do not explicitly model inter-frame information. Temporal convolutions, like C3D~\cite{c3d} extend the success of 2D convolutions in image understanding to include 3D convolutions across frames that are adjacent in time. Finally, recurrent convolutional networks~\cite{SrivastavaMS15, NgHVVMT15, DonahueHRVGSD17} have been used for learning video representations, most often by applying a 2D CNN as a feature extractor on the individual frames and using a recurrent network on the CNN features to model temporal information. Some of the latest methods actually combine the two-stream approach with the temporal convolution approach like the I3D architecture~\cite{i3d} or the R2+1D~\cite{r2plus1d}. In this paper we will focus on this hybrid family of methods, that include two streams and temporal convolutions, since they will allow us to experiment and compare the impact of testing and training on temporal data when using flow vs using images, and using frame-based models vs using temporal convolutions.

\noindent \textbf{Video Datasets}.
Progress in video understanding has been greatly propelled by the growth of video datasets in data scale and richness of taxonomy.
Two of the first datasets in this area were UCF-101~\cite{UCF101}, HMDB-51~\cite{hmdb}, which have been widely studied since the early 2010's. These are single-label videos of human actions in 101 classes and 51 classes, respectively. The first large-scale dataset to appear was Sports-1M~\cite{sports1m}, which provides 1M videos of sport actions in 487 classes. Although the labels are predicted by tagging algorithms and thus noisy, Sports-1M has consistently improved the quality of models pre-trained on it. In the last few years, researchers have pushed video classification to more challenging scenarios by using human annotation to collect large-scale datasets with different characteristics. For example, the Something-Something~\cite{some-some} dataset contains categories of generic actions independent of objects, with the goal of obtaining models that reason over physical aspects of actions and scenes as opposed to high-level action concepts. Kinetics~\cite{kinetics} includes 400 categories of human actions in different environments, combining the success case of UCF101 with large-scale. Unlike past datasets that focus on third person videos, Charades~\cite{charades} 
%argue that many of these datasets suffer from the bias of videos found on the internet, which are often unusual and eventful, and therefore not a faithful representation of the visual world. Thus the Charades dataset 
focuses on egocentric first-person videos that happen around the house. These datasets are often created by first deciding on a taxonomy of classes and then looking for videos containing instances of such classes, or recording them~\cite{monfortmoments}. This process may bias the distribution of videos based on the search engine, and thus SOA~\cite{soa} proposes a dataset where the videos are chosen first, and then labeled using the taxonomy of classes that naturally arises. While this is a more expensive labeling process, it yields videos often with multiple labels, and datasets with less biased label taxonomy. 
%It also combines labels of scenes and objects.
Other recent datasets also extend the types of video tasks to action localization(\eg THUMOS 14~\cite{jiang2014thumos}, ActivityNet~\cite{caba2015activitynet}, AVA~\cite{ava}), video object detection (\eg VLOG~\cite{vlog}), and video prediction (\eg Epic-kitchens~\cite{epic}).
%Other recent datasets include AVA~\cite{ava}, which focuses on atomic actions, and contains spatio-temporal labels, pushing towards action localization; VLOG~\cite{vlog} which focuses on object detection on a similar setting to SOA, where the set of videos is fixed first and then the classes are discovered.
%and Moments~\cite{monfortmoments}, which focuses on 3-second videos at a very large scale, combining audio and video. 
%Most recently,  Epic-kitchens~\cite{epic} focuses on ego-centric views of cooking videos, and provides rich labeling including temporal localization, captions, object detection, etc. 
% This wealth of new datasets reveals that the video community is still trying to explore and understand what constitutes a good task and good labels for video understanding. 
Despite the tremendous efforts of dataset collection, it is not well studied yet what action classes substantially require temporal modeling to recognize as opposed to still appearance modeling.

% \begin{figure}[t]
% \centering
% \includegraphics[width=\linewidth]{figures/example_canvas.png}
% % \includegraphics[width=0.45\linewidth]{figures/example_applying_lipstick.png}
% % \includegraphics[width=0.45\linewidth]{figures/example_brush_painting_soa.png}
% \caption{{\bf Examples of frames of a shuffled video in our experiment.} The canvas setting allows us to show the frames of the video without temporal information.}
% \label{fig:annotation_samples}
% \end{figure}

\noindent \textbf{Temporal Ordering and Frame Shuffling}. Temporal ordering in the video is arguably important for video understanding. A rich set of methods have explicitly exploited it for video recognition, including ranking pooling~\cite{fernando2017rank, cherian2017generalized}, and RNN-based methods~\cite{donahue2015long, SrivastavaMS15}. 
%Frame shuffling has been used before as an effective way to probe the  of temporal modeling methods. 
Furthermore, Misra et al.~\cite{shuffle2016} use it as a self-supervised task to learn better temporal models, where a network is trained to predict the original ordering of a shuffled set of frames. Other recent work has explored the effect of shuffling to measure the robustness of a network to temporal disarray~\cite{gcpr18, Huang2018WhatMA}, observing that even networks using temporal convolutions can be surprisingly insensitive to shuffling frames within clips. Finally, perceptual studies also have investigated the effect of shuffling in space and time~\cite{Koenderink12}. %\lt{removed the following as it seems to conflict with our motivation}, suggesting that human observers are able to recognize actions after some amount of shuffling has taken place. 
In contrast, in this work, we conduct a human perception study, and use frame shuffling to identify action classes where the action recognition performance of human is substantially compromised by the randomly shuffled temporal ordering.

% What actions are needed to understand action recognition? ~\cite{sigurdsson2017actions}

%\paragraph{Perception studies. } (Koendrick, Bei)

\section{The Temporal Dataset}
\label{sec:motion_cat}

% \begin{table*}[t]
% \begin{center}
% \small
% \begin{tabular}{l l l r}
% Dataset & Group & Classes & \hspace{-1cm} $\nabla$ Accuracy \\ \hline \hline
% \multirow{3}{*}{Kinetics} & Dancing & Belly dancing, breakdancing, capoeira, cheerleading, etc. & 35.0\% \\
% & Snow + Ice & Ski jumping, ski crosscountry, ski slalom, etc. & 24.68\% \\
% & Head + Mouth & Headbanging, headbutting, sneezing, sniffing, yawning, etc. & 22.02\% \\ \hline
% \multirow{3}{*}{\specialcell{Something-\\Something}} & Moving camera & Moving away with camera, approaching with camera, etc. & 51.38\% \\
% & Multi-object & Moving sth. closer to sth., moving sth. away from sth., etc. & 33.33\% \\ %\rule{0pt}{2.5ex}\\
% & Showing & Showing sth. to the camera, showing sth. next to sth., etc. & 27.22\%\\ %\rule{0pt}{2.5ex}\\ 
% \hline
% \end{tabular}
% \vspace{.1cm}
% \caption{{\bf Semantic groups where temporal information matters.} Action classes where scenes, objects and poses do not contain much information are part of the Temporal Dataset. }
% \label{tab:top_kinetics_groups}
% %\vspace{-0.1cm}
% \end{center}
% \end{table*}

In this section we describe the process to discover the categories for which temporal information is important, which will be part of what we call the Temporal Dataset. We first describe the perceptual test in detail and then discuss our findings. 

\subsection{Perceptual Test to Discover Temporal Action Classes}

\paragraph{Motivation.} How can we measure if temporal information is necessary to identify an action in a video? 
% For some of the categories a thought experiment is enough, and we can imagine that a single frame clearly contains the relevant information (\eg smiling, walking) or clearly does not (\eg moving left, moving right). Many categories however, may be somewhere in the middle. For example, the category ``swinging" may be easy to guess from a single frame if it depicts a child on a swing suspended mid-air, but might be more difficult if we see a monkey on a tree, who may be hanging, jumping down, or climbing. In this case we would say that temporal information is necessary ``at times". Our goal is to measure how often temporal information is necessary to identify each category. We will then select the classes that very frequently require temporal information to be recognized. %\lt{Hmmm, is this an accurate description? } \ls{Lorenzo, is the issue that we don't label the entire dataset?}
We could, for example, compare the accuracy of a method using the entire video as input to the accuracy using a single frame. However, going from the full video to a single frame eliminates not only the temporal information, but also some appearance information, since we would be excluding most of the frames. Further, recent studies have shown~\cite{Huang2018WhatMA} that some frames make recognition easier than others, raising a more complicated ``frame selection problem". Instead we maintain all image information and removes all temporal information by shuffling frames in time, and using the resulting video as input. 
We could then choose an action recognition network and observe the performance drop with and without temporal shuffling. However, this approach would tie the selection of temporal classes to a specific computational model. This is risky for two reasons. First, it conflates the inherent complexity of identifying a class and the temporal information. Second, we do not know the architecture's ability to represent temporal information in the first place. %For example, although networks with temporal convolutions are designed to capture temporal information, their strong performance in action recognition may be due more to their large learning capacity rather than their ability to capture motion. 
Our intent here is not to find classes where a specific model does poorly, but to define a ``meta-dataset" of temporal classes that will spur innovation in temporal modeling. 
Based on these arguments we propose to show the shuffled videos to human subjects and to define temporal relevance of a class based on the drop in accuracy in human recognition performance when the video frames are shuffled in time. 

%It would for example not be obvious whether one should choose an architecture like C3D~\cite{c3d}, with larger capacity, or one like R(2+1)D~\cite{r2plus1d} that tends to achieve higher accuracy, or one that uses an LSTM. In addition it would be difficult to control for the complexity of each class (e.g.: how difficult it is for a network to recognize), which may vary greatly for current networks. Instead, we show the shuffled videos to human subjects. \ls{not so happy with this paragraph}

%\ls{not sure if this needs more explanation...} \lt{Yes, the motivation for this design choice should be expanded to emphasize that this makes the selection independent of a specific computational model, etc.} 

%IMPORTANT for RESEARCH -- Measure without and without shuffling and the gap is the ability to model temporal info?? 

\paragraph{Stimuli.}

We empirically found that watching videos where frames are shuffled in time can be discomforting. It is hard for humans to attend to so much change in the scene. Even if we slow down the frame rate there is a fundamental problem: we are simply replacing the true temporal information with random temporal information; the human perceptual system will still attempt to connect the frames in time. For this reason we render the video on a canvas containing space for two frames side by side. At any time we black out one half (either the left or the right half) of the canvas and render the current frame in the other half. %(as shown in Fig.~\ref{fig:annotation_samples}). 
The next frame is then shown in the half that was previously blacked out. This left-right alternation eliminates any perception of motion (real or random), producing a slide-show effect where image information is retained but temporal information is removed. Video examples are included in the supplementary material.

\paragraph{Task.} Action recognition networks are typically trained for multi-class classification, where given an input video, they classify the action in it. Ideally we would like to expose human annotators to the exact same task. However, this is difficult in practice, since the number of classes is in the order of hundreds. Annotators would have a hard time remembering all the classes in the taxonomy. Another option is to ask the binary question ``is X action happening in this video?". The problem there is that we only have positive examples, and generating compelling negative examples is not trivial. Instead, for each video we show as possible answers only classes within a semantic group. For example, if a video has the label ``barbecueing", then we will show as possible answers other cooking categories, which includes classes like ``making a sandwich", since they are more likely to be visually similar. \footnote{These semantic groups are provided in the Kinetics dataset, but for the Something-something dataset we design them manually, using the verb as guidance.} This restricts the options that we give to the annotator, from the order of hundreds to 5-15 options. While this may be considered making the task easier, in practice it would be rare that categories across groups would be accidentally mistaken for each other.

\paragraph{Datasets.} 
We discover temporal classes from the most relevant action classification datasets. We use Kinetics~\cite{kinetics} (400 classes, 280K videos), for being the most widely used set. We also use Something-Something~\cite{some-some} (174 classes, 100K videos) because it was specifically designed to be action-centered and independent of the objects appearing in the scene which makes it particularly relevant to our study. %\lt{removed this sentence to avoid pissing reviewers involved in moments.} Another potential candidate for our analysis is Moments in Time~\cite{monfortmoments}, which unfortunately we are unable to use because of licensing reasons.

\paragraph{Selection of Temporal Classes.} We show between 15 and 30 videos per class to different human annotators. % We aggregate all annotations within each group, which has between 5 and 15 classes.
For each class $i$ we compute the average accuracy of the annotators when they see the videos shuffled $\textrm{Acc}_s(i)$ and for the control $\textrm{Acc}_c(i)$, when the videos are not shuffled. We select the groups where the control accuracy is larger than $70\%$ (\eg, the classes names are sufficiently clear) and the drop in accuracy $\textrm{Acc}_c(i) -\textrm{Acc}_s(i)$ is larger than a threshold ($40\%$ in Kinetics, and $60\%$ in Something-something, since performance drops have different statistics). We also include those action classes that have been the source of confusion more than $20\%$ of the times, since they are necessary to actually observe the confusion (\eg plugging and unplugging, opening and closing, etc).   

% \paragraph{Selection of Static Classes.} . \ls{todo}

 %As discussed above, we can analogously compute measures of accuracy-drop for individual classes. ~\ls{revise this methodology}%We decide to include entire groups instead of single classes because the decision of the annotators is dependent on the adversarial classes. For example, ``opening" might only be confused with ``closing", so if we were to only include one of them the final classification task may be easier.    

% In this section we compare human accuracy with and without frame shuffling information, in order to discover the categories for which temporal information is most important in all three datasets. 

%

% \ls{TODO: how many samples. experiment on repeatability}
% \ls{specific description of selection process }
% \ls{mention why using groups instead of classes}

%-- Temporal downsampling 
%-- Alternating sides 
%-- Groups 
%-- Control experiment 
%-- Measuring success 

% \subsection{Additional Classes}
% -- Non-rigid motion 
% -- Mannequin Challenge 
% -- 
%\subsection{What is Motion For?}
\subsection{What Actions Actually Require Temporal Information?}
% from python notebook: Accuracy per group and per class 

\begin{figure}[t]
\begin{floatrow}
\ffigbox{
\includegraphics[width=0.49\linewidth]{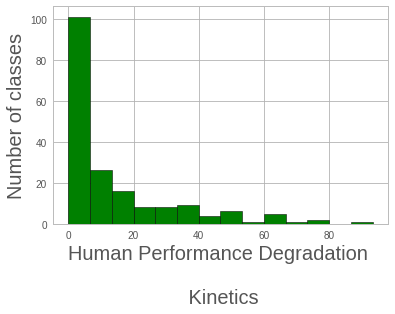}
\includegraphics[width=0.49\linewidth]{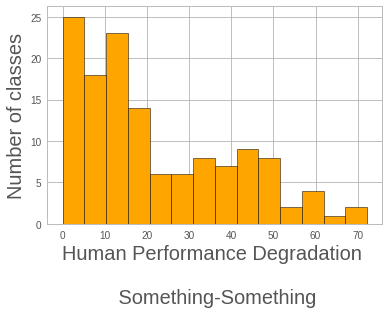}
% \ffigbox{%
%   \rule{3cm}{3cm}%
% }
}{%
  \caption{{\bf Histogram of performance drop (\%).} Something-Something contains more categories where temporal information is necessary. }%
}
\capbtabbox{%
  \begin{tabular}{l l r}
        Class &  Confused with & \hspace{-1cm} $\nabla$Acc (\%) \\ \hline \hline
        Ski jump. & Ski, Ski slal. & 73.33 \\
        Play cards & Shuffle cards & 66.66 \\
        Ski crossc. & Ski jump., Ski slal. & 60.00 \\  
        Drop kick & Wrestle, High kick & 60.00 \\
        Dance macar. & Breakdance, Tap  & 60.00 \\
        Shake head & Sneeze, Headbang & 53.33 \\
        Sneeze & Yawn, Blow nose & 46.66 \\ \hline
        \label{fig:dataset_analysis}
    \end{tabular}
}{%
  \caption{{\bf What is motion for? } Classes where human recognition accuracy drops most when temporal information is removed.}%
}
\end{floatrow}
\end{figure}

% \begin{figure*}[t]
% \centering
% \includegraphics[width=0.32\linewidth]{figures/histogram_kinetics_motion_categories_2.png}
% %\includegraphics[width=0.32\linewidth]{figures/histogram_soa_motion_categories_2.png}
% \includegraphics[width=0.32\linewidth]{figures/histogram_something_motion_categories_2.png}
% \caption{{\bf Histogram of human performance degradation (\%) when frames are shuffled.} Many categories in current datasets can be identified from still pictures, while some require temporal analysis to be recognized. Out of the three datasets considered here, the Something-Something dataset has the largest fraction of categories for which temporal information is necessary for accurate recognition. }%\lt{regenerate these histograms to have percentages on the x-axis (i.e. multiply by 100). Also I'd suggest to label x-axis as ``human accuracy degradation (\%) caused by frames reshuffling''}}
% \label{fig:histogram_perf_decay}
% \end{figure*}

% So what are the actions that require temporal information to be recognized? 

In total, we discover 50 temporal classes, where 32 come from the Kinetics datasets and 18 from Something-something. We call this subset the Temporal Dataset. The total amount of videos is 35045, where 32081 are for training and 2964 for testing. 

In Table 1 we show the actions where accuracy drops 
most when we remove temporal information, by group and by class, respectively.
As expected, temporal information is more important when the scene or objects in are not discriminative for the action recognition task, for example in videos showing different forms of dancing, or different skiing styles. These are classes where the background, clothing, and objects tend to be fairly constant across categories or uninformative for action recognition. We also observe that facial actions are particularly sensitive to frame shuffling, making actions like yawning and sneezing difficult to tell apart without temporal information. %Finally, higher-level actions that tell a story (e.g., auctioning or giving an award) also benefit from temporal information for recognition.  

Figure 3 shows the histogram of performance drop of all classes after temporal shuffling. The proportion of classes that require temporal analysis is small for both datasets. 
%These histograms are obtained by computing the difference in recognition accuracy with and without frame reshuffling for each class. %\lt{is the previous sentence accurate?}
Something-Something shows a larger proportion of categories requiring temporal information for discrimination. This is not surprising since this dataset was designed precisely to avoid strong correlations between object categories and action classes. Perhaps what may be more surprising is that still many of the classes in such dataset can be correctly identified from shuffled frames. %\lt{where is this more than half coming from? are you thresholding the drop in accuracy?} 
This highlights the need for the proposed methodology: it is difficult to design a dataset of actions that require strong temporal modeling for accurate recognition, while it is much easier to discover temporal classes using our simple perceptual test.

\section{Analysis of the Temporal Dataset}
%At this point we have a mechanism to reliably compute human accuracy with and without temporal information for each given action class. We argue that this can serve as a proxy to determine which actions require temporal information to be identified, and group them together into a 

%We now examine the properties of the classes that require temporal information to be recognized, which we call {\em temporal categories}. We aim to address the following questions: Which are these categories? What do they have in common? How do machines perform on these categories in comparison to the other ones? 

% A main motivation for this work is to find classes to serve as a benchmark for training and assessing truly temporal models. Thus, in this section we use the temporal categories to evaluate existing methods and observe that performance in these categories indeed favors models with the ability to capture temporal information. We also explore the use of the temporal categories to improve temporal representations via finetuning. 
\subsection{Experimental Details}
\label{subsec:exp_details}
% \begin{figure*}[t]
% \centering
% \includegraphics[width=\linewidth]{figures/sneezing_vs_crying.png}
% \caption{{\bf Examples of videos where it is difficult to tell classes apart from still images.} In these videos it is hard to assess whether the action is yawning, sneezing, or crying. }
% \label{fig:sneezing_vs_crying}
% \end{figure*}

\paragraph{Datasets.} We use two sets of classes in our experiments: {\em Temporal-50}, which is the set of 50 temporal action classes where performance decreases most as in Sec.~\ref{sec:motion_cat}. 
And {\em Static-50}, the 50 classes where human accuracy decreases least, also as measured in Sec.~\ref{sec:motion_cat}. The number of videos in this set is comparable to that of Temporal-50. While there is a larger portion of classes where accuracy decreases a negligible amount, we choose 50 at random to keep the number of classes and videos comparable to Temporal-50. 
% {\em Random-50:} This is a set of 50 classes chosen at random from the union of the two datasets. 
%\vg{Do these control classes remain fixed for all the experiments. If yes can we get how many motion/static classes are in this set?} 
%\end{itemize}
%\end{itemize}
Note that both sets contain the same proportion of classes from Kinetics, and Something-something, to factor out the dataset bias component. 

\paragraph{Architectures.} We choose a suite of architectures that are top-performing and most widely used. They include some that use temporal information (\eg through optical flow or temporal convolution) and others that do not. In particular, we use: R2D which is a frame-based model, that uses the ResNet~\cite{resnet} architecture with 18 layers, and it is trained with a cross-entropy loss; %Both the number of layers and the loss are consistent throughout the experiments. 
R3D~\cite{c3d} which is similar to R2D, but instead of processing each frame separately it includes convolutions in the temporal dimension;  
I3D~\cite{i3d} which is also a 3D architecture, that extends inception to use convolutions in the temporal dimension; and R(2+1)D which is a recent architecture~\cite{r2plus1d} that achieves state-of-the-art accuracy. The main difference compared to R3D is that R(2+1)D factors the 3D convolutions into 2D filters in space and 1D filters in time. This separability helps the optimization, making this the strongest of the models we use. 
%\lt{The following two sentences are my wild guess, please correct as needed.} For clip-based models we compute video-level accuracy by averaging predictions over 10 random clips taken from the video. For the frame-based model (R2D) we compute clip-level (video-level) accuracy by averaging frame-predictions over each clip (video).
%\ls{ I3D... ?}

\paragraph{Training and Evaluation Details. } Unless otherwise specified, all training experiments are performed with the same parameters: 75 epochs, from which 10 are warm-ups and the remaining 65 follow a cosine function, where the base learning rate is 0.0025 and gamma is 0.0001, on 4 GPUs at a time. The batch size is 16. We use multiple crops of size 112$\times$112. All models are trained from scratch. Unless otherwise specified, the reported evaluation of models is done using a clip size of 16 frames. We report only the video-level top-1 accuracy in the main paper for brevity.%, since we are interested in measuring relative performance. %We will include the clip-level and top-5 accuracy in the supplemental material for completeness. 

\subsection{Computational Properties of the Temporal Dataset}
\label{subsec:properties}

\begin{figure}[t]
  \begin{subfigure}[b]{0.45\textwidth}
        \centering
        \includegraphics[width=0.95\textwidth]{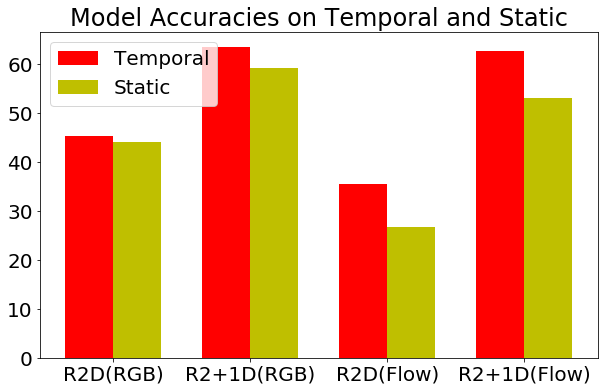}
        \caption{{Absolute Accuracy (\%).} }
    \end{subfigure}\hfill
    \begin{subfigure}[b]{0.45\textwidth}
        \centering
        \includegraphics[width=0.95\textwidth]{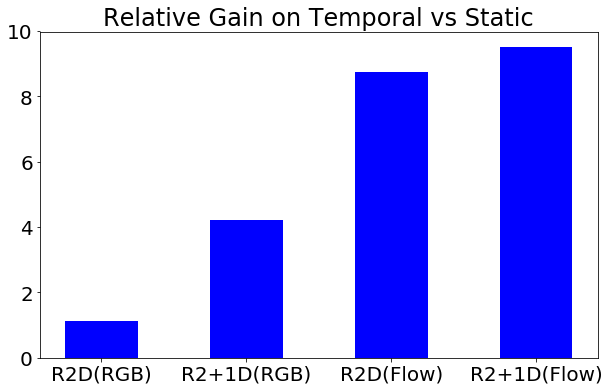}
        \caption{{Relative Accuracy (\%).} }
    \end{subfigure}
    \caption{{\bf The better a network represents temporal information, the better it performs on the Temporal Dataset.} R2D(RGB) has similar performance in the Temporal and Static Datasets (a). However, R2+1D(RGB) which adds temporal convolutions and shows better performance on the Temporal than the Static set. This is consistent with our hypothesis, that the Temporal Dataset helps identify a network's ability to model temporal information. This pattern is consistent when we switch from RGB to Flow. This difference is clearly observed in (b). }
    %\vspace{-0.15in}
    \label{fig:correlation}
\end{figure}

In this section we perform a series of computational sanity checks, that help validate the nature of the Temporal and Static datasets. This is, that the Temporal Dataset actually captures the classes where temporal information matters, and the Static Dataset captures those where temporal information is redundant. We argue that a randomly chosen set of classes is very unlikely to satisfy all our sanity checks.

% How can we cross-check that the Temporal Dataset actually captures the classes where temporal information matters and the Static Dataset captures those where temporal information is redundant? Ideally we would have an alternative method for discovering temporal and static classes that leads to similar sets. But finding such alternative method is not trivial. Instead, we perform a series of sanity checks, and hypothesize that each of them brings additional certainty about the nature of the Temporal and Static Datasets, and that it would be very unlikely that two random sets of classes would satisfy all these sanity checks. 

\paragraph{Temporal and Static sets are statistically different.}

The first sanity check is to make sure that the behavior of networks is different in the Temporal and Static Datasets in a statistically significant manner. Given a network $N$, we compute the per-class accuracy of the network $\textrm{Acc}_N(i)$ of all classes $i$. We now compute all accuracies of the classes in the Temporal Dataset $T$, as $A = \{\textrm{Acc}_N(i) \; | \; i \in T\}$. We also compute the accuracies in the Static Dataset $S$, which is the set $B = \{\textrm{Acc}_N(i) \; | \; i \in S\}$. We use the Kolmogorov-Smirnov test to measure if $A$ and $B$ are sampled from the same distribution, for multiple networks $N$ (R2+1D and R2D). We observe that the null hypothesis can be rejected with $p < 0.1$, suggesting that the behavior of the networks on the datasets $T$ and $S$ are sampled from different distributions. As a control experiment, we randomly sample two sets $T'$ and $S'$, and observe that $p > 0.4$. We conclude that the behavior of networks is statistically different in the Temporal and Static datasets. 

\paragraph{A network's ability to capture temporal information is correlated with its accuracy in the Temporal Dataset.} Recall that one of our goals with the creation of the Temporal Dataset is to use it as a benchmark to measure a network's ability to capture temporal information. In this sanity check we use pairs of networks where we do know their relative abilities to capture temporal information. In particular we use the pair R2D and R2+1D, which are identical except for the fact that R2D is frame-based and R2+1D includes temporal convolutions. Thus it is safe to say that R2+1D has a stronger ability to represent temporal information than R2D. We also use the pair RGB and flow. If the Temporal Dataset actually contains videos where temporal information is necessary, we should observe that R2+1D has an advantage over R2D on the Temporal Dataset, and so does using flow as input over using RGB. Of course, R2+1D has larger capacity, and it will typically always outperform R2D. We use the accuracy on the Static Dataset to control for the capacity of the networks, and measure {\em the difference between the accuracy in the Temporal Dataset and the Static Dataset}. 

Results are shown in Fig.~\ref{fig:correlation}. The accuracy of the R2D network using RGB as input is similar in the Temporal and the Static Datasets (Fig.~\ref{fig:correlation}(a)). However, when we use the R2+1D architecture, which adds temporal convolutions, the network does better on the Temporal than the Static set. This is consistent with our hypothesis, that networks that model temporal information better perform better in the Temporal Dataset. Similarly, we observe better performance on the Temporal Dataset when using flow as input, than when using RGB as input. This difference is more clearly observed in Fig.~\ref{fig:correlation}(b), which shows the correlation between the networks' ability to capture temporal information, and its gain in the Temporal Dataset, as expected. %We also observe this pattern in Something-something and Kinetics when we analyzed them independently. 
At the same time, we do not observe this pattern in the control experiment of sampling subsets of classes at random. With this we conclude that {\em the relative gain in the Temporal Dataset compared to the Static Dataset is a good measure for evaluating a network's temporal capabilities}.

%\paragraph{Impact of Shuffling.}
%asergsegges

%\input{human_vs_machine}
%\ls{for now removing the human vs machine}

%\subsection{Flow Accuracy and temporal Categories}
%\subsection{How Can we Discover Temporal Categories without Humans in the Loop}
%\ls{shuffling experiments?}
%\subsection{Optical Flow Accuracy and Action Accuracy}
\section{Temporal Dataset as a Benchmark}
%As mentioned above, we want to discover the action classes that require temporal modeling.  

\begin{table}[t]
\centering
    \begin{tabular}{l r r r r r r r r}
    Network & \specialcell{R2D\\(RGB)} & \specialcell{R2D\\(Flow)} & \specialcell{R2+1D\\(RGB)} & \specialcell{R2+1D\\(Flow)} & 
    \specialcell{I3D\\(RGB)} & 
    \specialcell{I3D\\(Flow)} &
    \specialcell{R3D\\(RGB)} & 
    \specialcell{R3D\\(Flow)}\\ \hline \hline
    \specialcell{Temporal\\Score ($\nabla$ \%)} & 1.13 & 8.74 & 4.21 & {\bf 9.50} & 3.60 & 6.29 & 5.19 & 7.36 \\ \hline
    \specialcell{Traditional\\Accuracy (\%)} &
    31.83 & 22.23 & 44.58 & 41.92 & {\bf 51.53} & 43.35 & 50.48 & 45.50 \\ \hline
    \end{tabular}
    \caption{{\bf Evaluation of popular action recognition networks. } The ranking of methods changes when we use the traditional score and when we use the proposed temporal score. %\ls{add Kinetics and Something-something}
    }
    \label{tab:benchmark}
\end{table}

In this section we propose a score to evaluate a network's ability to capture temporal information based on the experiments from Sec.~\ref{subsec:properties}. We use this score to rank some of the most recent and widely used networks. %Based on our observations in Sec.~\ref{subsec:properties}, 
In particular, we propose computing the relative gain of the network on the temporal classes, compared to static classes: 
\begin{equation}
    \textrm{Score}(N) =  \overline{\textrm{Acc}}_N(T) - \overline{\textrm{Acc}}_N(S),
\end{equation}
where $\textrm{Acc}_N(T)=\{\textrm{Acc}_N(i) \mid i \in T \}$ and $\textrm{Acc}
_N(S) =\{\textrm{Acc}_N(i) \mid i \in S\}$.
Results of evaluating a suite of popular methods are shown in Table~\ref{tab:benchmark}. 

We make several observations. First, using optical flow shows to be much more useful. This reveals that the perception that optical flow is not useful in recent datasets is a more a product of the task (i.e., solving Kinetics) than the true nature of the visual world. Second, we observe that the top-performing method (I3D) according to the traditional accuracy, scores rather poorly in the temporal score. This is quite possibly a direct consequence of the training process using inflation based on image filters learned from ImageNet. While this process is useful for action recognition, it biases the network to excel in static classes. Since R2+1D and R3D are not pre-trained on Imagenet, their overall performance on the datasets is lower, but they do shine on the temporal classes.

\section{Effect of Training on the Temporal Dataset}

We also explore the effect of training using the Temporal Dataset. In particular, we are interested in observing the behavior of the training in the absence of discriminative image information. In principle, this should force the network to learn stronger temporal features, which may help generalization. %We test two specific cases: transfer learning, and sensitivity to shuffling temporal structure. 
For the experiments we use the R2+1D arquitecure with RGB as input. 

\begin{figure}[t]
  \begin{subfigure}[b]{0.45\textwidth}
        \centering
        \includegraphics[width=0.95\textwidth]{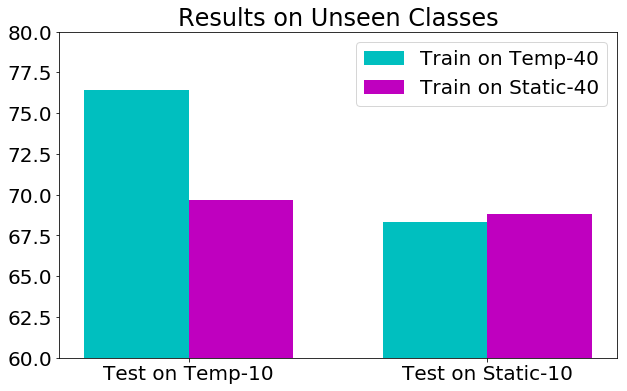}
        \caption{{Transfer learning.}}
    \end{subfigure}\hfill
    \begin{subfigure}[b]{0.45\textwidth}
        \centering
        \includegraphics[width=0.95\textwidth]{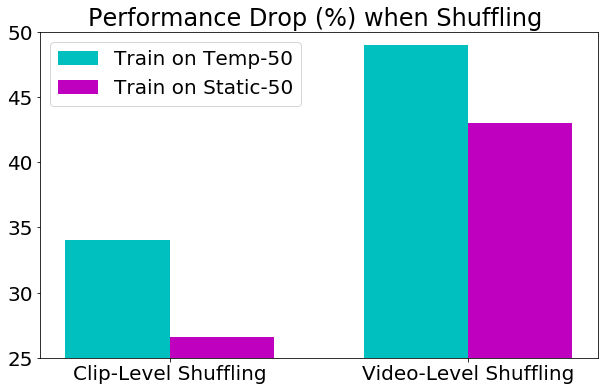}
        \caption{{Shuffling at test time.} }
    \end{subfigure}
    \caption{{\bf Effect from training on the Temporal Dataset.} (a) Learning from temporal classes generalizes better to unseen classes than training on static classes. (b) Shuffling at test time has a much stronger effect on networks trained on temporal classes.}
    \label{fig:training}
    %\vspace{-0.1in}
\end{figure}

\paragraph{Training on the Temporal Dataset avoids image bias and generalizes better. } In this experiment we measure the ability of a network to generalize to unseen classes when the network has been trained on temporal classes, and on static classes. We take each of the two datasets, which contain 50 classes, and we train on 40 of them, and leave the other 10 as unseen classes to test. After training a network on 40 classes, we freeze all weights except for the last layer, which we change to map to 10 classes, and we finetune it. Results are shown in Fig.~\ref{fig:training} (a). We would expect that training on temporal classes would be better for testing on temporal classes, and that training on static would be better for testing on static. However, we find that testing on static shows on par performance for both training processes. We hypothesize that in the absence of discriminative image cues, the network cannot ``cheat" and is forced to learn better temporal information, which is useful both for static and temporal classes. More importantly, we find that training on temporal performs much better testing on temporal, which shows the need of temporal data to learn good temporal features. 

% \begin{table*}[t]
% \begin{center}
% \small
% \begin{tabular}{l l r}
% Training & Testing & Accuracy \\ \hline \hline
% \multirow{2}{*}{Temp-40} & Temp-10 &  0\\
% & {Static-10} & 0 \\ \hline
% \end{tabular}
% \vspace{.1cm}
% \caption{{\bf Groups that require temporal information for disambiguation, from Kinetics and Something-Something. %Some of the groups of classes require temporal ordering to be identified. 
% }}
% \label{tab:top_kinetics_groups}
% %\vspace{-0.1cm}
% \end{center}
% \end{table*}
\paragraph{Training on the Temporal Dataset produces features more sensitive to temporal changes.} Shuffling frames at test time has been noted to not harm performance dramatically~\cite{gcpr18}. This observation reveals that features are not very sensitive to changes in temporal structure. Does training on the Temporal Dataset produce more temporally-aware features? We measure the effect of shuffling frames on networks trained on static and temporal classes. Results are shown in Fig.~\ref{fig:training} (b). We observe that the performance of a network trained on temporal classes drops much more than one trained on static classes. This is consistent both when shuffling at the video level and the clip level. We argue that training on temporal classes leads to larger sensitivity to temporal ordering, and therefore stronger temporal features. 

\section{Conclusion}

We have presented a methodology to discover the relevance of temporal information in action classes based on human perception. We use it to identify categories that contain more temporal information and we called them temporal classes. We use this set of classes to benchmark current video models, revealing a network's ability to model temporal information. We also used these classes to for training and observe that boosts accuracy on the temporal categories, even on unseen classes. We hope that the proposed ``meta-dataset" of temporal categories will help guide future research in temporal modeling for better video understanding.\footnote{Answers: ``sneezing"  (left), and ``crying" (right).}

\section*{Appendix}

{\bf Temporal classes indices}: \\
\\
From Kinetics: [30, 34, 41, 45, 63, 75, 105, 147, 148, 152, 173, 177, 206, 222, 228, 230, 235, 277, 295, 301, 302, 307, 308, 309, 310, 322, 325, 349, 357, 358, 376, 395]
\\
From Something-something: [1, 6, 10, 12, 26, 34, 39, 45, 47, 56, 60, 63, 67, 70, 77, 81, 95, 150]\\
\\
{\bf Static classes indices}: \\
\\
From Kinetics: [18, 20, 23, 26, 53, 57, 80, 88, 91, 95, 99, 113, 126, 139, 181, 182, 183, 186, 200, 201, 210, 218, 226, 227, 255, 268, 298, 354, 366, 388, 397, 398]
\\
From Something-something: [9, 15, 42, 50, 58, 61, 63, 71, 78, 90, 93, 106, 109, 114, 121, 130, 138, 165]\\
\\
{\bf Random sets indices}: \\
\\
From Kinetics: [7, 8, 27, 30, 41, 44, 65, 73, 80, 86, 112, 113, 118, 125, 137, 146, 158, 189, 203, 204, 205, 206, 208, 210, 228, 231, 254, 355, 362, 377, 385, 399]
\\
From Something-something: [0, 2, 5, 18, 27, 31, 38, 58, 68, 103, 105, 110, 113, 132, 135, 144, 147, 164]\\
\\
%%%
{\bf Temporal classes names}: \\
\\
From Kinetics: [bouncing on trampoline, breakdancing, busking, cartwheeling, cleaning shoes, country line dancing, drop kicking, gymnastics tumbling, hammer throw, high kick, jumpstyle dancing, kitesurfing, parasailing, playing cards, playing cymbals, playing drums, playing ice hockey, robot dancing, shining shoes, shuffling cards, side kick, ski jumping, skiing (not slalom or crosscountry), skiing crosscountry, skiing slalom, snowboarding, somersaulting, tap dancing, throwing ball, throwing discus, vault, wrestling]\\
\\
From Something-something: [Turning something upside down, Approaching something with your camera, Moving something away from the camera, Moving away from something with your camera, Moving something towards the camera, Lifting something with something on it, Moving something away from something, Moving something closer to something, Uncovering something, Pretending to turn something upside down, Covering something with something, Lifting up one end of something, then letting it drop down, Lifting something up completely without letting it drop down, Moving something and something closer to each other, Moving something and something away from each other, Lifting something up completely, then letting it drop down, Stuffing something into something, Moving something and something so they collide with each other]\\
\\
{\bf Static classes names}: \\
\\
From Kinetics: [belly dancing, bending back, blasting sand, blowing nose, changing wheel, clapping, curling hair, deadlifting, dining, doing aerobics, dribbling basketball, eating doughnuts, filling eyebrows, getting a tattoo, laying bricks, long jump, lunge, making bed, moving furniture, mowing lawn, peeling apples, playing badminton, playing controller, playing cricket, pull ups, riding camel, shot put, testifying, trimming trees, waxing eyebrows, yawning, yoga]\\
\\
From Something-something: [Folding something, Turning the camera upwards while filming something, Holding something next to something, Pouring something into something, Pretending to throw something, Squeezing something, Lifting up one end of something, then letting it drop down, Holding something in front of something, Touching (without moving) part of something, Lifting up one end of something without letting it drop down, Showing something next to something, Poking something so that it falls over, Wiping something off of something, Scooping something up with something, Letting something roll down a slanted surface, Sprinkling something onto something, Pushing something so it spins, Twisting (wringing) something wet until water comes out]\\
\\
{\bf Random sets names}: \\
\\
From Kinetics: [arranging flowers, assembling computer, blowing out candles, bouncing on trampoline, busking, carrying baby, cleaning windows, cooking sausages, curling hair, dancing gangnam style, eating chips, eating doughnuts, egg hunting, feeding goats, gargling, grooming horse, hugging, making snowman, opening bottle, opening present, paragliding, parasailing, passing American football (in game), peeling apples, playing cymbals, playing flute, presenting weather forecast, texting, tossing salad, waiting in line, watering plants, zumba]\\
\\
From Something-something: [Holding something, Turning the camera left while filming something, Opening something, Moving something up, Showing a photo of something to the camera, Something falling like a feather or paper, Tearing something just a little bit, Pretending to throw something, Attaching something to something, Moving something across a surface without it falling down, Dropping something next to something, Moving something across a surface until it falls down, Pulling something out of something, Pushing something off of something, Pretending or failing to wipe something off of something, Something colliding with something and both are being deflected, Putting something onto something else that cannot support it so it falls down, Pretending to spread air onto something]\\
\\
% % # unseen labels: pick 10 classes from temporal, in which 6 from kinetics, 4 from something
% Seen classes: 
% Temporal: From Kinetics: [34, 41, 45, 63, 75, 147, 148, 152, 173, 177, 222, 228, 230, 235, 277, 301, 302, 307, 308, 309, 322, 325, 349, 357, 358, 376], From Something-something: [6, 10, 12, 26, 39, 45, 47, 56, 60, 67, 70, 77, 81, 95]}\\
% \\
% Static: From Kinetics: [20, 23, 26, 53, 57, 88, 91, 95, 99, 113, 139, 181, 182, 183, 186, 201, 210, 218, 226, 227, 268, 298, 354, 366, 388, 397], Something-something: [15, 42, 50, 58, 63, 71, 78, 90, 93, 109, 114, 121, 130, 138]
% \\
% Random: {'kinetics': [8, 27, 30, 41, 44, 73, 80, 86, 112, 113, 125, 137, 146, 158, 189, 204, 205, 206, 208, 210, 231, 254, 355, 362, 377, 385], 'something': [2, 5, 18, 27, 38, 58, 68, 103, 105, 113, 132, 135, 144, 147]}}\\
% \\
% Unseen classes: {'temporal': {'kinetics': [30, 105, 206, 295, 310, 395], 'something': [1, 34, 63, 150]}, 'static': {'kinetics': [18, 80, 126, 200, 255, 398], 'something': [9, 61, 106, 165]}, 'random': {'kinetics': [7, 65, 118, 203, 228, 399], 'something': [0, 31, 110, 164]}}
% \clearpage

%\section*{References}

{\small
\bibliographystyle{ieee}
\bibliography{egbib}
}

\end{document}